\pdfoutput=1

\documentclass[11pt]{article}

\usepackage[]{EMNLP2023}

\usepackage{times}
\usepackage{latexsym}

\usepackage[T1]{fontenc}

\usepackage[utf8]{inputenc}

\usepackage{microtype}

\usepackage{inconsolata}

\usepackage{graphicx}
\usepackage{amsmath}
\usepackage{amsfonts}
\usepackage{booktabs}
\usepackage{xfrac}
\usepackage{multirow}

\usepackage{tikz}
\usetikzlibrary{fadings}
\usetikzlibrary{patterns}
\usetikzlibrary{shadows.blur}
\usetikzlibrary{shapes}
\usetikzlibrary{shapes.geometric,calc,tikzmark}

\ExplSyntaxOn
\NewDocumentCommand \liaison { O{0.5mm} O{20} m m }
  {
    \begin{tikzpicture}[remember~picture, overlay]
      \draw ([yshift=-#1]#3.south) to[bend~right=#2] ([yshift=-#1]#4.south);
    \end{tikzpicture}
  }
\ExplSyntaxOff

\title{Evaluating Transformer's Ability to Learn\\ Mildly Context-Sensitive Languages}

\author{Shunjie Wang\thanks{\ \ This work extends the author's master's thesis \citep{wang2021evaluating} done at University of Washington.} \\
  Independent Researcher \\
  \texttt{shunjiewang@hotmail.com} \\\And
  Shane Steinert-Threlkeld \\
  University of Washington \\
  \texttt{shanest@uw.edu} \\}

\begin{document}
\maketitle
\begin{abstract}
Despite the fact that Transformers perform well in NLP tasks, recent studies suggest that self-attention is theoretically limited in learning even some regular and context-free languages. These findings motivated us to think about their implications in modeling natural language, which is hypothesized to be mildly context-sensitive. We test the Transformer's ability to learn mildly context-sensitive languages of varying complexities, and find that they generalize well to unseen in-distribution data, but their ability to extrapolate to longer strings is worse than that of LSTMs. Our analyses show that the learned self-attention patterns and representations modeled dependency relations and demonstrated counting behavior, which may have helped the models solve the languages.
\end{abstract}

\section{Introduction}

Transformers \citep{vaswani2017} have demonstrated well-versed language processing capabilities and enabled a wide range of exciting NLP applications ever since its inception. However, 
\citet{hahn2020} shows that hard self-attention Transformers, as well as soft attention under some assumptions, fail at modeling regular languages with periodicity as well as hieararchical context-free languages eventually when presented with long enough sequences.

These theoretical limitations have since sparked the interest of the formal language community. A variety of formal languages, as well as formal models of computation such as circuits, counter automata, and predicate logic, have been studied to characterize the expressiveness of the architecture.

When it comes to probing an architecture's linguistic adequacy, a particular class of formal languages and formalisms naturally comes into sight: the \textit{mildly context-sensitive} class \citep{joshi_1985, kallmeyer2010}, the formal complexity class hypothesized to have the necessary expressive power for natural language.

\begin{figure}
    \centering
    \includegraphics[width=\columnwidth]{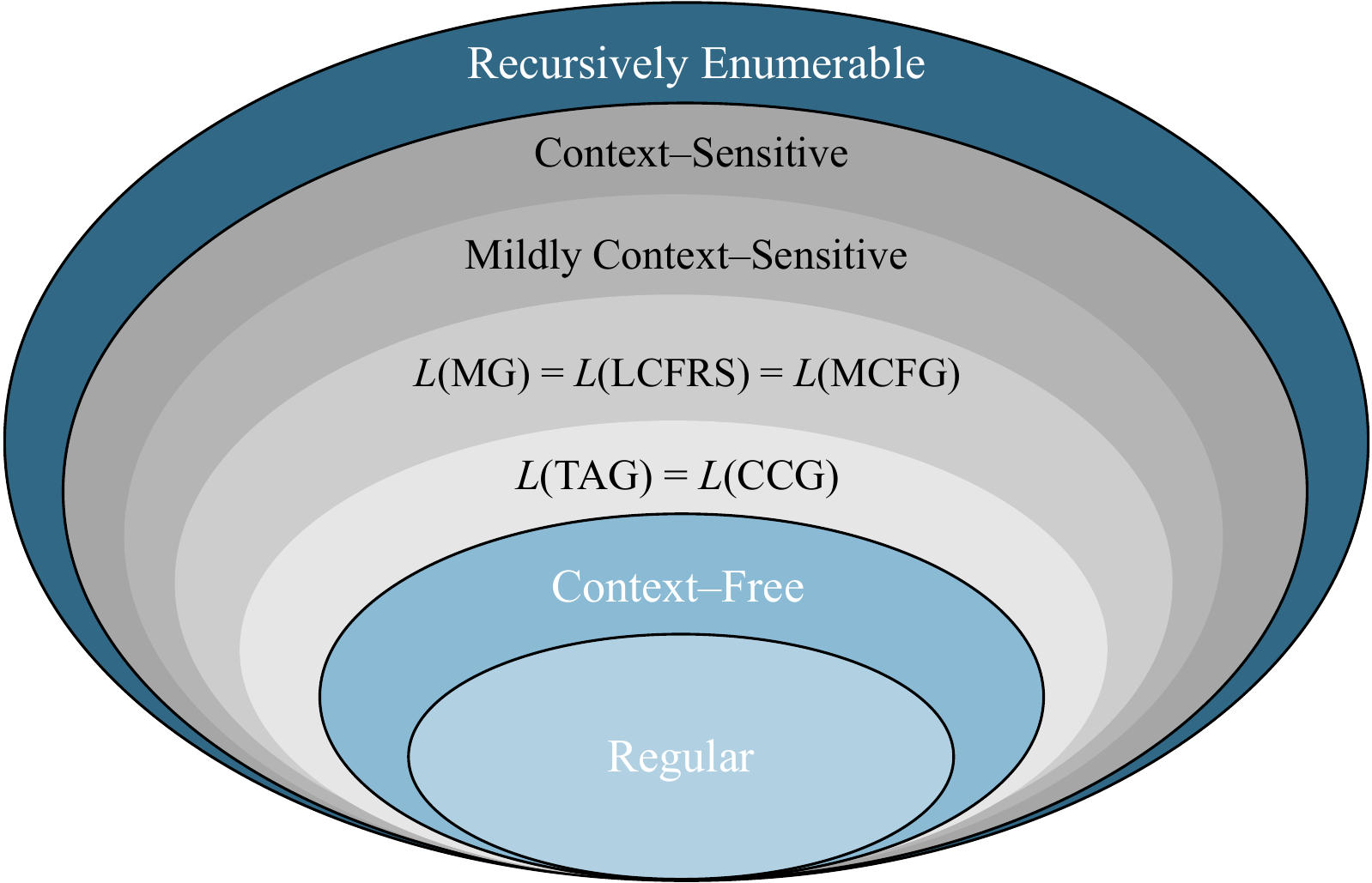}
    \caption{How certain MCSGs fit on the Chomsky hierarchy of languages in terms of their weak generative capacities \citep{Stabler2011}: $\text{CFL} \subset L(\text{TAG}) = L(\text{CCG}) \subset L(\text{MG}) = L(\text{LCFRS}) = L(\text{MCFG})\subset \text{CSL}$. No formalism generates the largest set of MCSLs.}
    \label{fig:CH}
\end{figure}
\begin{figure*}
    \centering
    \includegraphics[width=\textwidth]{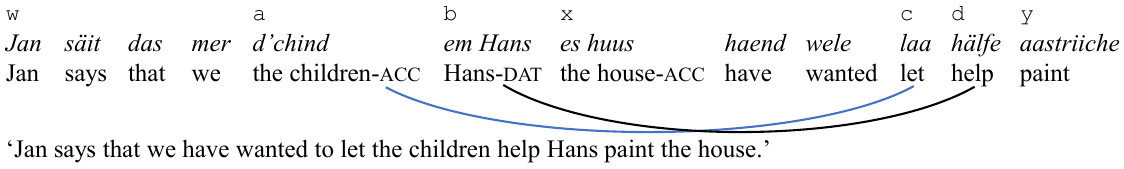}
    \caption{Swiss German subordinate clauses allow $n$ accusative NPs before $m$ dative NPs, followed by $n$ corresponding accusative object taking verbs before $m$ corresponding dative object taking verbs. \citet{Shieber1985} defined a homomorphism for Swiss German such that when intersecting with regular $\mathtt{w}\mathtt{a}^*\mathtt{b}^*\mathtt{x}\mathtt{c}^*\mathtt{d}^*\mathtt{y}\mathtt{z}$ yields non-context-free $\mathtt{w}\mathtt{a}^n\mathtt{b}^m\mathtt{x}\mathtt{c}^n\mathtt{d}^m\mathtt{y}\mathtt{z}$, which contradicts CFL's closure property under intersection with regular languages.}
    \label{fig:my_label}
\end{figure*}

This motivates us to study the Transformer's ability to learn a variety of linguistically significant, mildly context-sensitive string languages of varying degrees of complexities. Specifically, we ask two research questions:
\begin{enumerate}
    \item How well do Transformers learn MCSLs from finite examples, both in terms of generalizing to in-distribution data, as well as extrapolating to strings longer than the ones seen during training?
    \item What kind of meaningful representations or patterns do the models learn?
\end{enumerate}

Our contributions include that we extend current empirical studies on formal language learning with Transformers to the mildly context-sensitive class, and find that they generalize well to unseen strings within the same length range as training data, but their ability to extrapolate is worse than that of LSTMs. We also present analyses that suggest self-attention learned symbol dependency relations, and the representations encoded count information for some complex languages, which may have been useful for solving languages in this class.

\section{Mildly Context-Sensitive Hypothesis}

In search of computational models that have adequate power to generate natural language sentences while also assigning meaningful structural descriptions like trees to them, \citet{chomsky1956, CHOMSKY1959137} defined context-sensitive grammars (CSG) and context-free grammars (CFG) as intermediate systems that lie between two extremities: the Turing machine which overgenerates and the finite-state automaton which undergenerates. A question that immediately follows the definitions is whether the CFG could serve as a computational model for natural language, which had been an open question for a few decades until it was settled by evidence such as Swiss German cross-serial dependency (Figure \ref{fig:my_label}; \citealp{Shieber1985}) and Bambara vocabulary \citep{Culy1985}, which demonstrated the existence of natural languages that are supra-context-free.

However, although more restricted than the Turing machine, the CSG is also undesired as it still has much more generative capacity than natural languages should ever need, and as a result of that, is hard to parse efficiently. This motivated Tree-Adjoining Grammars (TAG; \citealp{joshi_1985}) and Combinatory Categorial Grammars (CCG; \citealp{Steedman1987}) among a few other weakly equivalent formalisms such that extend the CFG with just enough additional descriptive power so that phenomena like Swiss German cross-serial dependency can be treated, while not using the full CSG thus parsing can still be efficient. The properties of these formalisms with such additional power roughly characterize a class of languages and grammars that \citet{joshi_1985} calls \textit{mildly context-sensitive languages/grammars} (MCSL/MCSG).

Another related line of weakly equivalent formalisms, such as Linear Context-Free Rewriting Systems (LCFRS; \citealp{vijay-shanker-etal-1987-characterizing}), Multiple Context-Free Grammars (MCFG; \citealp{SEKI1991191}), and Minimalist Grammars (MG; \citealp{stabler1997}), further extend their expressive power beyond that of the TAG, as motivated by more complex phenomena like German scrambling \citep{becker-etal-1991-long}. While no single grammar formalism generates the largest possible set of MCSLs that satisfy the formal characterization in \citet{kallmeyer2010}, these are the closest approximations we have. Such differences in expressiveness formed a subhierarchy within this class (Figure \ref{fig:CH}), and the languages recognizable by TAGs and MGs and their respective equivalents, denoted as $L(\text{TAG})$ and $L(\text{MG})$, are therefore the language complexity classes that we examine in this work.
          
Therefore, one hypothesis for the complexity of natural language is that it is mildly context-sensitive (henthforth MCS). There have been some challenges citing linguistic data requiring more power beyond MCS \citep{radzinski-1991-chinese, 10.1007/BFb0052165, kobele2006generating}, but the validity of these claims remains controversial \citep{10.1162/ling.2004.35.4.683, clark2012beyond}, or no consensus has been reached on the need for more power \citep{graf2021mg}. Thus, while acknowledging that whether the MCS hypothesis is true remains an open question, it is a reasonably good hypothesis that allows us to analyze natural languages meaningfully.

\section{Related Work}
Regarding the expressiveness of the Transformer, \citet{perez2018on, JMLR:v22:20-302} established the Turing-completeness of the hard-attention Transformer. \citet{bhattamishra-etal-2020-computational} proves the Turing-completeness of soft attention by showing that they can simulate RNNs. However, these results assumed arbitrary precision for weights and activations, had certain departures from the original architecture, and made the proofs through their unique task definitions.

In a practical case of formal language learning from finite examples, the Transformer's ability is known to be limited, even in the regular language class. Empirically, it was shown that Transformers of different self-attention variants have limited abilities to learn certain star-free languages, as well as non-star-free, periodic regular languages \citep{hahn2020, bhattamishra-etal-2020-ability}, although the latter may still be recognized theoretically with a simple modification to the architecture \citep{chiang-cholak-2022-overcoming}.

As for context-free languages with hieararchical structural analyses such as Dyck-$n$, \citet{ebrahimi-etal-2020-self} empirically demonstrated one Transformer encoder setup in which such languages may be learned and observed stack-like behavior in self-attention patterns. \citet{yao-etal-2021-self} proved and empirically showed that self-attention can learn Dyck-$n$ with a bounded depth, although the boundedness reduces the CFL to regular. Additional empirical work on Dyck-$n$ include \citet{bernardy-etal-2021-Transformer, wen2023interpretability}, among others.

Besides language recognition guided by the Chomsky hierarchy, another line of research investigates other alternative formal languages, such as counter-recognizable languages \citep{bhattamishra-etal-2020-ability} and first-order logic \citep{merrill2023Transformers, chiang2023tighter}, to characterize the expressiveness of Transformers.

This work introduces MCSGs into the empirical explorations through assessing the ability of Transformers in certain learning settings to learn a variety of string languages recognizable by MCSGs of different complexities, which have not yet been studied as a whole like languages in other classes. Occasionally, studies on the Transformer's learning ability worked with data that conveniently fall into this class, including a few counter languages that are also TAG-recognizable \citep{bhattamishra-etal-2020-ability}, discontinuities in Dutch \citep{kogkalidis-wijnholds-2022-discontinuous}, reduplication \citep{deletang2023neural}, as well as a crossed parentheses language inspired by crossing dependency\footnote{However, their construction is not based on the language on which \citet{Shieber1985} based his argument, and a CFG analysis might exist for their string language.} \citep{papadimitriou2023pretrain}. This work complements these and other aforementioned related work by presenting a systematic evaluation guided by basic MCSL constructions and the subhierarchy within the class, as well as comparing each of the basic constructions against a less and a more complex counterparts.

\section{Methodology}

\subsection{Task Setups}

Our experiments use the original soft-attention Transformer with sinusoidal positional encoding (henthforth PE) as defined in \citet{vaswani2017}, and the architecture we use is based on a Transformer with just self-attention and feedforward sublayers, and depending on the task, we may use either unidirectional or bidirectional attention. We avoid using dropout as it could negatively impact performance since we are working with simple abstract formal languages. For each experiment, we also train an LSTM \citep{lstm} baseline for comparison. The implementations\footnote{The code for the experiments is available at: \\ \url{https://github.com/shunjiewang/mcsl-transformer}} for both the Transformer and the LSTM are from PyTorch \citep{pytorch}.

\begin{table*}
\centering
\addtolength{\tabcolsep}{+4.3pt}
\begin{tabular}{lcccc}
\toprule
& & \multicolumn{3}{c}{Mildly Context-Sensitive}
\\
\cmidrule{3-5}
& CFL & $L(\text{TAG})$ & \multicolumn{2}{c}{$L(\text{MG})=L(\text{MCFG})$}\\
& \textsc{less complex} & \textsc{canonical} & \textsc{more complex} & \textsc{scramble}\\
\midrule
Copying & $ww^\mathcal{R}$ & $ww$ & $www$ & \begin{tabular}{@{}c@{}} \ \\ \  \end{tabular}\\
Crossing Dependency & $\tikzmarknode{A}{\mathtt{a}^n}\tikzmarknode{B}{\mathtt{b}^m}\tikzmarknode{C}{\mathtt{c}^m}\tikzmarknode{D}{\mathtt{d}^n}$ \liaison{A}{D} \liaison{B}{C} & $\tikzmarknode{A}{\mathtt{a}^n}\tikzmarknode{B}{\mathtt{b}^m}\tikzmarknode{C}{\mathtt{c}^n}\tikzmarknode{D}{\mathtt{d}^m}$ \liaison{A}{C} \liaison{B}{D} & \begin{tabular}{@{}c@{}} \ \\ \  \end{tabular} & $O_2$\\
Multiple Agreements & $\mathtt{a}^n\mathtt{b}^n$ & \begin{tabular}{@{}c@{}} $\mathtt{a}^n\mathtt{b}^n\mathtt{c}^n$\\ $\mathtt{a}^n\mathtt{b}^n\mathtt{c}^n\mathtt{d}^n$\end{tabular} & $\mathtt{a}^n\mathtt{b}^n\mathtt{c}^n\mathtt{d}^n\mathtt{e}^n$ & MIX\\
\bottomrule
\end{tabular}
\caption{\label{tab:languages}
Languages this work studies, organized by complexities and basic MCSL constructions each represents or resembles.
}
\end{table*}

We use one of the following two established tasks for each of the languages depending on which better enables learning for the data. We further elaborate on the reasoning for the choice of task for each language in Appendix \ref{sec:details}.

\paragraph{Binary Classification (Bidirectional Attention)}
Following \citet{weiss-etal-2018-practical}, a model $g$ is said to recognize a formal language $L\in\Sigma^*$ if $f(g(w)) = 1$ for all and only strings $w \in L$. In our case, $g$ is a Transformer encoder, and $g(w)$ is the representation of a positive or negative example $w$, which is averaged from each symbol's encoder output for all symbols in $w$. $f$ is a fully connected linear layer that maps the pooled representation to a real number, which is then passed through the sigmoid function to obtain the class label 0 or 1 using a threshold of 0.5. The loss is the BCE loss between the prediction and the target label.

\paragraph{Next Character Prediction (Unidirectional Attention)}
For languages in which training with positive and negative examples is ineffective because too few examples are available or the set of possible negative examples is too large, we use this task, which only requires positive examples.
Given a valid prefix of a string in a language at each timestep, the model is tasked to predict the next set of acceptable symbols, or predicts \texttt{[EOS]} if the prefix is already in the language. To do this, the Transformer outputs for each symbol in the string are passed in parallel through a linear layer and then the sigmoid function using a threshold of 0.5 to obtain $k$-hot vectors of dimension $|\Sigma\cup\{\texttt{[EOS]}\}|$, where each dimension represents whether the symbol is in the set of next characters at the timestep. We consider the prediction for a string to be correct if all predicted $k$-hot vectors are correct. The loss is the individual symbol's BCE loss between the predicted and the target $k$-hot vectors summed and then averaged. A look-ahead mask is applied to prevent self-attention from attending to later positions, which indirectly offers positional information \citep{Irie2019, bhattamishra-etal-2020-ability, haviv-etal-2022-Transformer}, thus making PE optional for the languages we study using this task, and makes the architecture in this task essentially a Transformer decoder.

\subsection{Data}
Following the categorization in \citet{ILIE199733}, we are interested in three basic constructions that should be contained in MCSLs:
\begin{enumerate}
    \item copying: $ww$
    \item crossing dependency: $\mathtt{a}^n\mathtt{b}^m\mathtt{c}^n\mathtt{d}^m$
    \item multiple agreements: $\mathtt{a}^n\mathtt{b}^n\mathtt{c}^n$
\end{enumerate}

All these languages are TAG-recognizable. We try to compare each of the three languages with a similar but less complex context-free language, as well as a similar but more complex MG-recognizable language. We also investigate two relevant scramble languages that are also felicitously MCFG-recognizable.

\subsubsection{Copying}
Copy language $\{ww\mid w\in\{\mathtt{a},\mathtt{b}\}^*\}$ is in $L(\text{TAG})$ \citep{joshi_1985}. Its context-free counterpart is the palindrome $\{ww^\mathcal{R}\mid w\in\{\mathtt{a},\mathtt{b}\}^*\}$, where $w^\mathcal{R}$ is the reverse of the string $w$. \citet{joshi_1985} indicates double copy language $www$ is not in $L(\text{TAG})$. However, any multiple copying $w^k$ is in $L(\text{MG})$ \citep{Jager2012}. Thus, we study $www$ as the simplest strictly MG-recognizable language for copying.

We use the binary classification setup for this family of languages. To generate the strings, we enumerate each possible $w$ in our chosen $|w|$ range and then duplicate $w$ to produce $ww^\mathcal{R}$, $ww$, and $www$. Negative examples are random strings sampled from $\{\mathtt{a}, \mathtt{b}\}^*$ in the same length range as the positive examples, but are not in the set of positive examples.

\subsubsection{Crossing Dependency}
Cross-serial dependency language $\mathtt{a}^n\mathtt{b}^m\mathtt{c}^n\mathtt{d}^m$ is in $L(\text{TAG})$ \citep{joshi_1985}. Its context-free counterpart is the nesting dependency language $\mathtt{a}^n\mathtt{b}^m\mathtt{c}^m\mathtt{d}^n$. We use the next character prediction task for these two languages because the potential set of negative examples is too large. 

Following \citet{gersschmidhuber2001}, to recognize nested $\mathtt{a}^n\mathtt{b}^m\mathtt{c}^m\mathtt{d}^n$, when the input is $\mathtt{a}$, the next valid character is $\mathtt{a}$ or $\mathtt{b}$. As soon as the input becomes $\mathtt{b}$, the value of $n$ is determined, then the next valid symbol is now $\mathtt{b}$ or $\mathtt{c}$. Once the input becomes $\mathtt{c}$, the value of $m$ is also determined, and the next characters are deterministic from this point on. Lastly, we output \texttt{[EOS]} as soon as the final symbol in the input is consumed. Following the notation in \citet{suzgun-etal-2019-evaluating}, we denote the above described input-target scheme as the following, where $\dashv$ denotes \texttt{[EOS]}:
\begin{equation*}
    \mathtt{a}^n\mathtt{b}^m\mathtt{c}^m\mathtt{d}^n \rightarrow (\sfrac{\mathtt{a}}{\mathtt{b}})^n(\sfrac{\mathtt{b}}{\mathtt{c}})^m\mathtt{c}^{m-1}\mathtt{d}^n\dashv
\end{equation*}

Trivially, this scheme can be generalized to the crossing dependency language:
\begin{equation*}
    \mathtt{a}^n\mathtt{b}^m\mathtt{c}^n\mathtt{d}^m \rightarrow (\sfrac{\mathtt{a}}{\mathtt{b}})^n(\sfrac{\mathtt{b}}{\mathtt{c}})^m\mathtt{c}^{n-1}\mathtt{d}^m\dashv
\end{equation*}

\begin{table}
\centering
\footnotesize
\addtolength{\tabcolsep}{-0.1pt}    
\begin{tabular}{llccc}
\toprule
          && \textsc{In-Distr.}   & OOD     & OOD  \\
\cmidrule{3-5}
          && $|w|\in [1,11]$  & $|w|=12$      & $|w|=13$  \\
\midrule
\multirow{2}{*}{$ww^\mathcal{R}$} &Transf.& $\textbf{99.5}_{\pm 0.3}$ & $\text{50.4}_{\pm 0.3}$  & $\text{50.2}_{\pm 0.1}$                \\
&LSTM & $\text{97.8}_{\pm 0.5}$ & $\textbf{96.0}_{\pm 0.7}$                   &$\textbf{96.0}_{\pm 0.8}$              \\
\midrule
\multirow{2}{*}{$ww$}  &Transf.& $\textbf{99.5}_{\pm 0.1}$                 &  $\text{51.3}_{\pm 0.3}$                &  $\text{50.2}_{\pm 0.0}$               \\
&LSTM & $\text{97.2}_{\pm 0.4}$                 & $\textbf{95.7}_{\pm 1.1}$              &  $\textbf{90.4}_{\pm 1.4}$               \\
 \midrule
 \multirow{2}{*}{$www$} &Transf.&$\textbf{99.5}_{\pm 0.2}$ & $\text{51.0}_{\pm 0.5}$                  & $\text{50.5}_{\pm 0.2}$  \\
 &LSTM & $\text{99.4}_{\pm 0.1}$                   & $\textbf{98.6}_{\pm 0.5}$                   & $\textbf{87.5}_{\pm 6.2}$                 \\
\bottomrule
\end{tabular}
\caption{\label{tab:copying}
Palindrome/Copy/2-Copy: Transformers surpass LSTMs for in-distribution tests but fall to random guesses for OOD (null accuracy for OOD is 50\%).
}
\end{table}

\subsubsection{Multiple Agreements}
Being simple counter languages, the multiple agreements family had previously been extensively studied. We complement the related work by adding an MG-recognizable language, as well as giving additional analyses on the learned pattern.

Both $\mathtt{a}^n\mathtt{b}^n\mathtt{c}^n$, $\mathtt{a}^n\mathtt{b}^n\mathtt{c}^n\mathtt{d}^n$ are TAG-recognizable. Their context-free counterpart is $\mathtt{a}^n\mathtt{b}^n$. \citet{joshi_1985} indicates $\mathtt{a}^n\mathtt{b}^n\mathtt{c}^n\mathtt{d}^n\mathtt{e}^n$ is not in $L(\text{TAG})$, but \citet{stabler1997} shows that it is in $L(\text{MG})$. Moreover, $\sigma_1^n,...,\sigma_k^n$ for any arbitrary $k$ is MG-recognizable \citep{Jager2012}. Thus, we study $\mathtt{a}^n\mathtt{b}^n\mathtt{c}^n\mathtt{d}^n\mathtt{e}^n$ as the simplest strictly MG-recognizable language in this family.

Since the dataset is very small as only one positive example is available for each $n$, we use the next character prediction task, which requires fewer examples and is also an established setup for learning these languages. \citet{gersschmidhuber2001, suzgun-etal-2019-evaluating} have proposed the input-target schemes for $\mathtt{a}^n\mathtt{b}^n$, $\mathtt{a}^n\mathtt{b}^n\mathtt{c}^n$, and $\mathtt{a}^n\mathtt{b}^n\mathtt{c}^n\mathtt{d}^n$:
\begin{align*}
    \mathtt{a}^n\mathtt{b}^n &\rightarrow (\sfrac{\mathtt{a}}{\mathtt{b}})^n\mathtt{b}^{n-1}\dashv\\
    \mathtt{a}^n\mathtt{b}^n\mathtt{c}^n &\rightarrow (\sfrac{\mathtt{a}}{\mathtt{b}})^n\mathtt{b}^{n-1}\mathtt{c}^n \dashv\\
    \mathtt{a}^n\mathtt{b}^n\mathtt{c}^n\mathtt{d}^n &\rightarrow (\sfrac{\mathtt{a}}{\mathtt{b}})^n\mathtt{b}^{n-1}\mathtt{c}^n\mathtt{d}^n \dashv
\end{align*}

Trivially, this scheme can be generalized to the MG-recognizable $\mathtt{a}^n\mathtt{b}^n\mathtt{c}^n\mathtt{d}^n\mathtt{e}^n$:
\begin{align*}
    \mathtt{a}^n\mathtt{b}^n\mathtt{c}^n\mathtt{d}^n\mathtt{e}^n \rightarrow (\sfrac{\mathtt{a}}{\mathtt{b}})^n\mathtt{b}^{n-1}\mathtt{c}^n\mathtt{d}^n\mathtt{e}^n \dashv
\end{align*}

That is, the next valid character is $\mathtt{a}$ or $\mathtt{b}$ as long as the input is $\mathtt{a}$, but once $\mathtt{b}$ occurs in the input, $n$ will be determined and the next characters will be deterministic from this point on.

\subsubsection{Scramble Languages}

\citet{joshi_1985} argued that MCSGs should only handle limited cross-serial dependency like the type in Dutch \citep{Bresnan1982} and Swiss German, but not as in $\text{MIX} = \{w\in\{\mathtt{a},\mathtt{b},\mathtt{c}\}^*\mid |w|_\mathtt{a} = |w|_\mathtt{b} = |w|_\mathtt{c}\}$, where $|w|_\sigma$ denotes the number of occurrences of symbol $\sigma$ in string $w$, that is, the language of strings with an equal number of $\mathtt{a}$'s, $\mathtt{b}$'s, and $\mathtt{c}$'s, but the symbols can occur in any order, and can thus be seen as scrambled $\mathtt{a}^n\mathtt{b}^n\mathtt{c}^n$.

\begin{figure}
    \centering
    \includegraphics[width=\columnwidth]{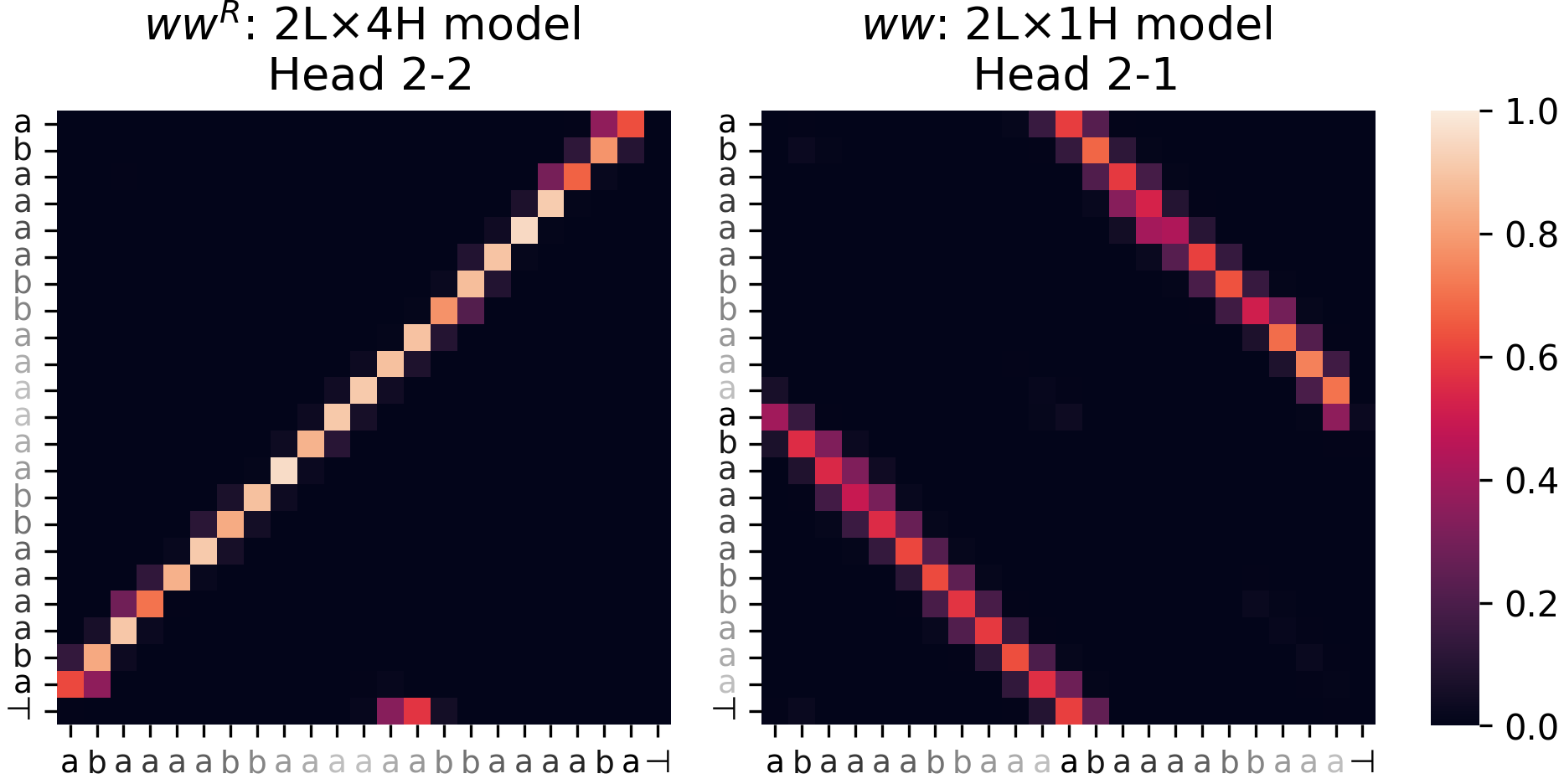}
    \caption{Anti-diagonal alignment for $ww^\mathcal{R}$, and forward and backward alignment for $ww$.}
    \label{fig: ww}
\end{figure}

MIX resembles an extreme case of free word order and is not recognizable by the TAG \citep{kanazawa-salvati-2012-mix}. However, it turned out that the language is in $L(\text{MCFG})$ \citep{Salvati2015}. A related language $O_2 = \{w\in\{\mathtt{a},\mathtt{b},\mathtt{c}, \mathtt{d}\}^*\mid |w|_\mathtt{a} = |w|_\mathtt{c} \wedge  |w|_\mathtt{b} = |w|_\mathtt{d}\}$, which can be seen as scrambled $\mathtt{a}^n\mathtt{b}^m\mathtt{c}^n\mathtt{d}^m$, is also in $L(\text{MCFG})$ \citep{Salvati2015}.

\begin{figure}
    \centering
    \includegraphics[width=\columnwidth]{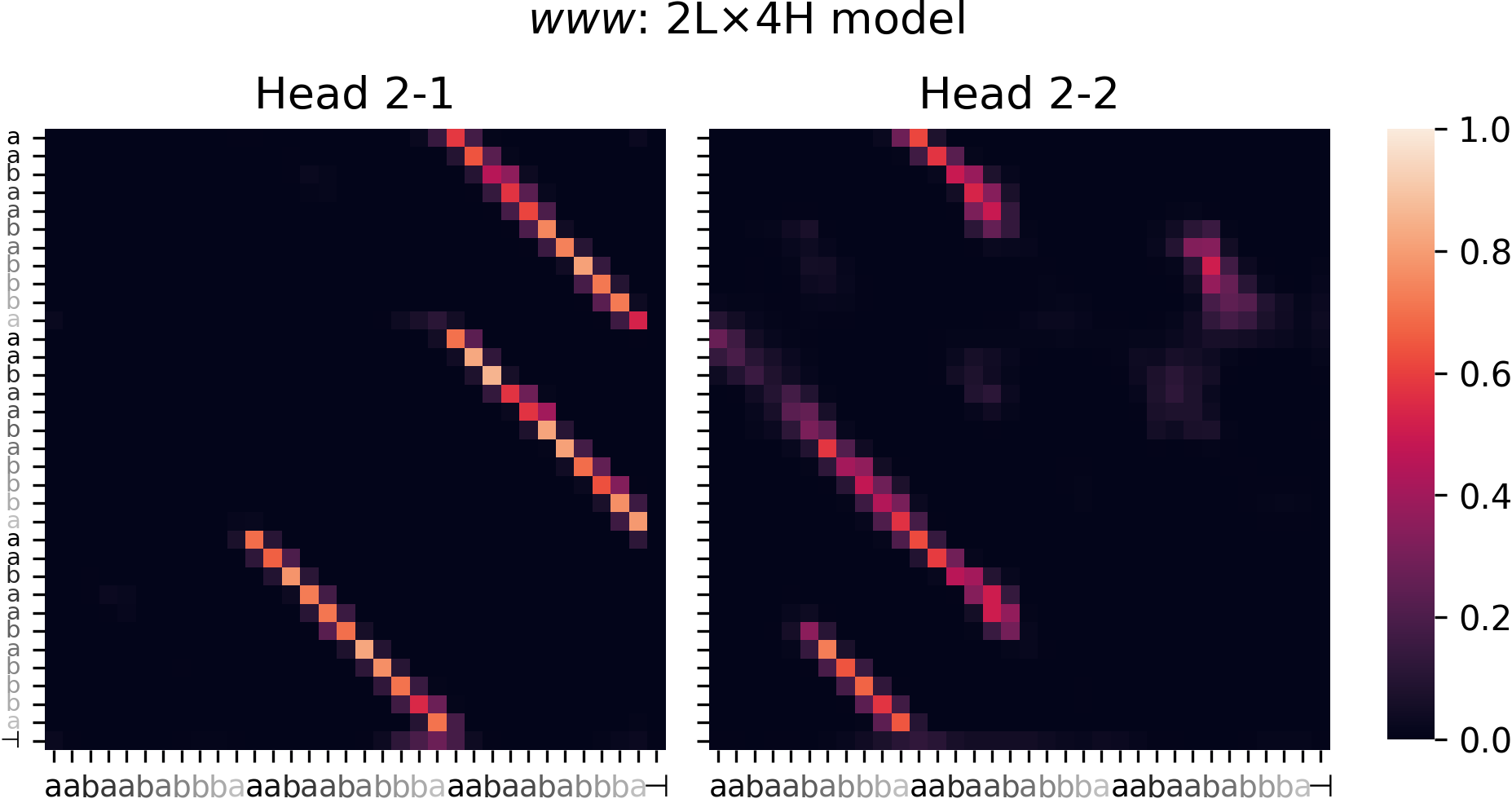}
    \caption{$www$ expects six alignments in total, usually distributed across heads, where half are better aligned, and the other half partially aligned.}
    \label{fig: www}
\end{figure}

We investigate the two scramble languages using the binary classification task as the model may benefit from seeing the whole string at once to directly model the occurrences of each symbol. For MIX, the positive examples exhaustively enumerate all permutations of $\mathtt{a}^n\mathtt{b}^n\mathtt{c}^n$ in the chosen $n$ range, and the negative examples enumerate the remaining strings in $\{\mathtt{a},\mathtt{b},\mathtt{c}\}^*$ within the same range to help the model better eliminate most wrong hypotheses. As for $O_2$, we enumerate permutations of $\mathtt{a}^n\mathtt{b}^m\mathtt{c}^n\mathtt{d}^m$ in the chosen $n,m$ range, and the negative examples are from the remaining strings in $\{\mathtt{a},\mathtt{b},\mathtt{c}, \mathtt{d}\}^*$ that are in the same range. Negative examples for both languages also include strings where $|w|_\sigma = 0$. The train-test split is performed over each sequence length separately rather than over the entire dataset, so strings of different lengths appear in all splits.

Table \ref{tab:languages} summarizes all languages studied in this work, and we also include detailed dataset statistics in Appendix \ref{sec:training}.

\section{Experiments}

Each model is evaluated on three sets: an in-distribution held-out test set, an out-of-distribution (henthforth OOD) set with strings longer than the ones seen during training, and a second OOD set with even longer strings. We report the mean and standard deviation of test accuracies in three runs with different random seeds.

We then visualize the heads from our best-performing runs that most clearly demonstrate highly interpretable patterns upon visual inspection, but note that all visualized patterns do recur across different configurations.

\subsection{Copying}
Our Transformer models learned $ww$ and the related $ww^\mathcal{R}, www$ with high accuracy and outperformed LSTMs in in-distribution tests. However, in the two OOD tests, only LSTMs were able to extrapolate, while the accuracies of the Transformers are close to random guesses (Table \ref{tab:copying}).

We identified certain heads that align the substrings in different diagonalities to measure the similarity of a string to itself (Figure \ref{fig: ww}). For $ww$, the gold alignment is to align the first $w$ against the second and vice versa. In the visualized run, we find that among all positive examples in the test set, 93.4\% of the time the highest query-key attention is on the gold alignment. For $ww^\mathcal{R}$, the gold alignment expects the head of a substring to attend to the tail of the other substring, thus resulting in an anti-diagonal pattern, and 94.8\% of the time the highest attention is on the gold alignment diagonals among all positive examples in the test set.

For $www$, a gold alignment requires each substring to attend to the other two, resulting in a total of six alignments, which we did get during training as shown in Figure \ref{fig: www}. The six alignments are distributed across heads in a multi-head model, where only three of the six are better aligned, while the other partial alignments appear to be auxiliary if were at all useful for inference. In the visualized model, the three clearer alignments in the first head on average match the gold alignment 86.1\% of the time over positive examples in the test set, while the accuracy is 39.0\% for the other three partial alignments.

\begin{table}
\centering
\footnotesize
\addtolength{\tabcolsep}{-2.9pt}    
\begin{tabular}{llccc}
\toprule
          && \textsc{In-Distr.}   & OOD     & OOD  \\
\cmidrule{3-5}
          && $n,m \in$     &$n$ or $m \in$     & $n$ or $m \in$ \\
          && ${[}1,50{]}$  & ${[}51,100{]}$    & ${[}101,150{]}$ \\
\midrule
&Tr.$+$PE& $\text{99.8}_{\pm 0.2}$                     &$\text{6.5}_{\pm 1.3}$                    &$\text{0.0}_{\pm 0.0}$                    \\
$\tikzmarknode{A}{\mathtt{a}^n}\tikzmarknode{B}{\mathtt{b}^m}\tikzmarknode{C}{\mathtt{c}^m}\tikzmarknode{D}{\mathtt{d}^n}$ \liaison{A}{D} \liaison{B}{C}&Tr.$-$PE & $\textbf{100.0}_{\pm 0.0}$  &$\text{98.0}_{\pm 0.3}$ &$\text{23.0}_{\pm 3.1}$\\
&LSTM & $\textbf{100.0}_{\pm 0.0}$                    & $\textbf{100.0}_{\pm 0.0}$                   & $\textbf{100.0}_{\pm 0.0}$                   \\
\midrule  
&Tr.$+$PE & $\textbf{100.0}_{\pm 0.0}$ & $\text{7.2}_{\pm 1.6}$& $\text{0.0}_{\pm 0.0}$\\
$\tikzmarknode{A}{\mathtt{a}^n}\tikzmarknode{B}{\mathtt{b}^m}\tikzmarknode{C}{\mathtt{c}^n}\tikzmarknode{D}{\mathtt{d}^m}$ \liaison{A}{C} \liaison{B}{D}&Tr.$-$PE & $\textbf{100.0}_{\pm 0.0}$ & $\text{92.3}_{\pm 1.2}$& $\text{27.0}_{\pm 14.0}$\\
&LSTM &$\textbf{100.0}_{\pm 0.0}$                   & $\textbf{99.2}_{\pm 1.3}$                   & $\textbf{81.3}_{\pm 12.3}$                    \\
\bottomrule
\end{tabular}
\caption{\label{tab:crossing}
Nesting/Crossing: not using sinusoidal PE helped with extrapolation. Note that the two OOD sets include both strings where only one of $n,m$ is OOD, e.g., $\mathtt{a}^1\mathtt{b}^{100}\mathtt{c}^1\mathtt{d}^{100}$, and strings where both $n,m$ are OOD, e.g., $\mathtt{a}^{51}\mathtt{b}^{100}\mathtt{c}^{51}\mathtt{d}^{100}$.
}
\end{table}

\begin{figure}
    \centering
    \includegraphics[width=\columnwidth]{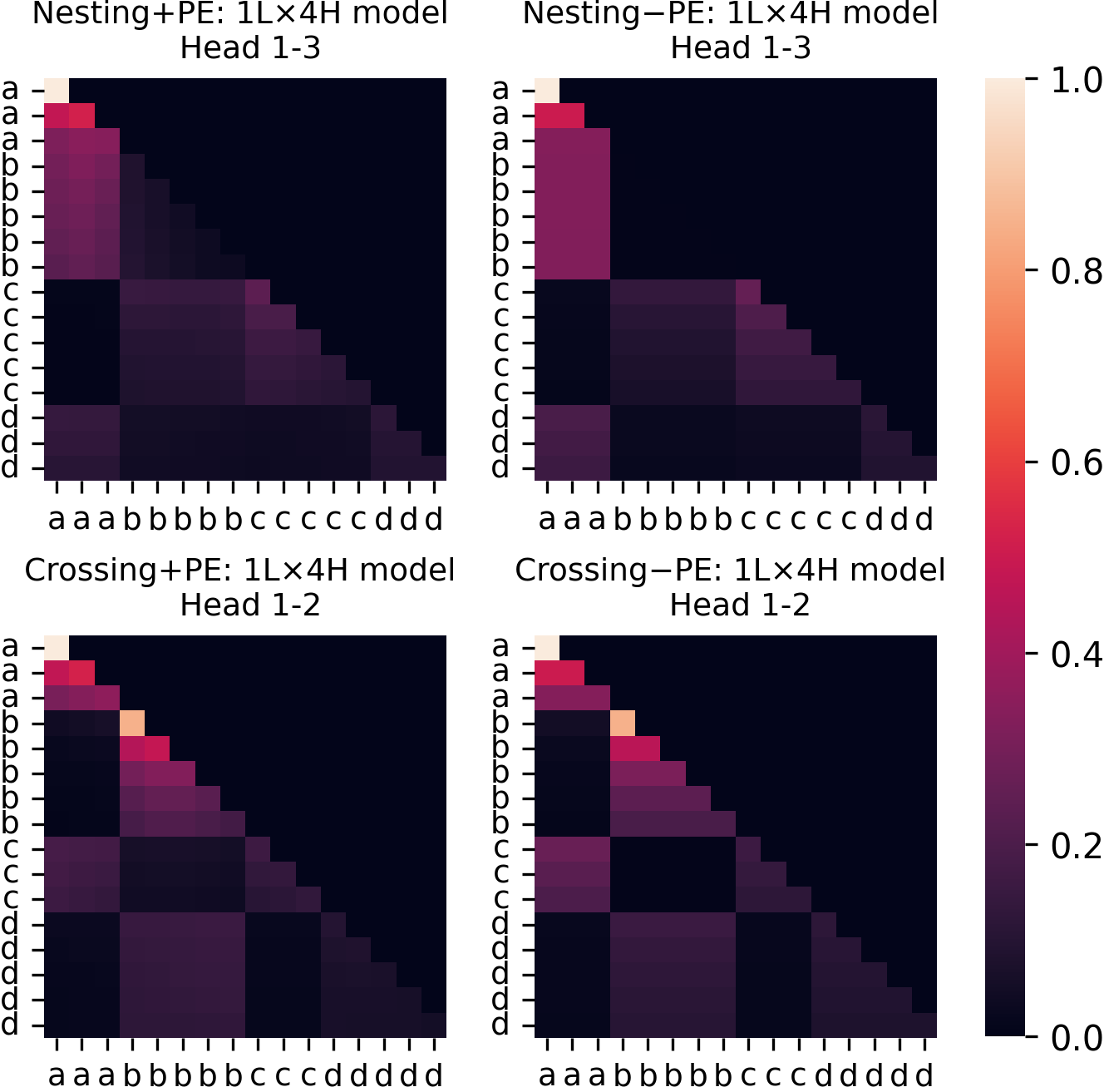}
    \caption{Crossing shows a checkerboard pattern as the result of correctly identifying pairwise dependents, while nesting has a similar pattern except for different symbol dependencies.}
    \label{fig: cross}
\end{figure}

\subsection{Crossing Dependency}
The in-distribution tests were solved almost perfectly by all three model setups, including a Transformer decoder setup where we remove PE and rely only on indirect positional information from the causal mask. For OOD tests, we evaluate models on strings where both $n, m$ are OOD, as well as strings in which only one of $n, m$ is OOD. We find that removing PE and relying only on causal mask helped with the Transformer's extrapolation, which is consistent with findings in \citet{bhattamishra-etal-2020-ability} on other languages. However, such ability is still worse than that of LSTMs (Table \ref{tab:crossing}).

Figure \ref{fig: cross} shows that the Transformer's attention formed a checkerboard pattern for recognizing crossing, as the result of each symbol in the query attends to every occurrence of itself and its dependent in the key but not to the non-dependents. As for nesting, the pattern is very similar except that the dependency relations are different. Models trained with or without sinusoidal PE end up learning very similar patterns, except that without PE, the attention from one query symbol to every occurrence of a key symbol is uniformly distributed, resulting in a stack of color bands on the attention map of the visualized head.

\begin{figure}
    \centering
    \includegraphics[width=\columnwidth]{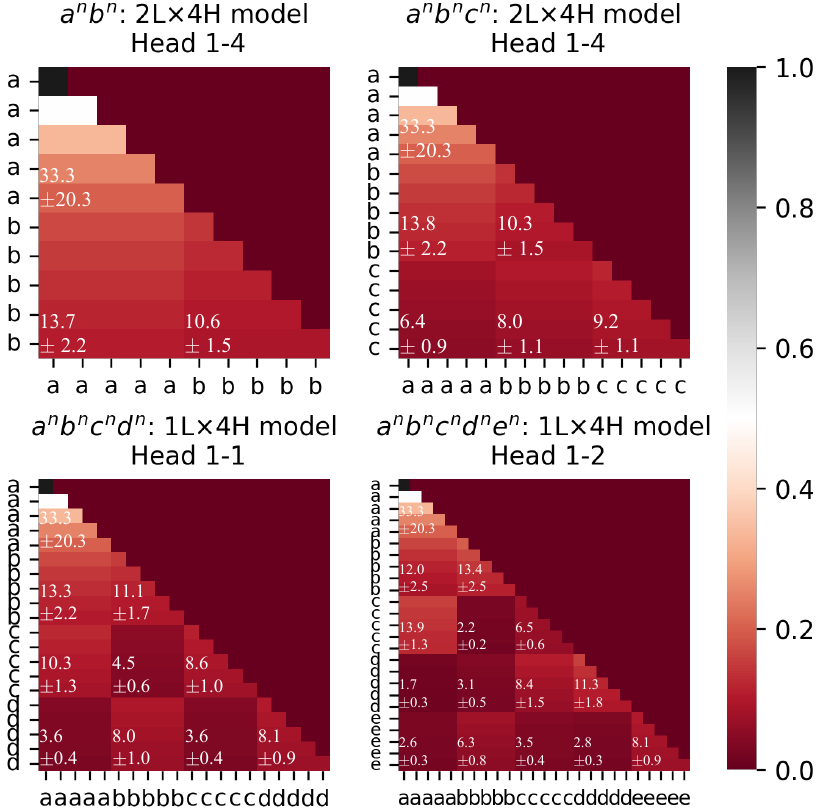}
    \caption{Every occurrence of a query symbol attends to every occurrence of a key symbol using similar attention values, thus low variance among the values. The query symbols also attend to different key symbols with different weights, thus resulting in a grid pattern.}
    \label{fig: multiple}
\end{figure}

The attention maps suggest that in the optimal case, each symbol in query identifies to which other symbol in key it is pairwise dependent, and then in the visible portion without look-ahead mask, distributes its attention to every occurrence of itself and the dependent, and gives zero attention to the other pair. We measure as an example how accurately the visualized head in the crossing model without PE has implemented this optimum, and we find that across all in-distribution test set datapoints, keys that expect 100\% of the attention weights from each symbol in query have received on average 93.0\% of the attention weights.

\subsection{Multiple Agreements}
We follow the established finding in \citet{bhattamishra-etal-2020-ability} and only consider the Transformer decoder without sinusoidal PE for these languages, as training with PE was ineffective in pilot experiments. Transformers without PE demonstrated the ability to extrapolate, although on average they are still not as good or consistent as LSTMs (Table \ref{tab:multiple}).

\begin{table}
\centering
\footnotesize
\addtolength{\tabcolsep}{-3pt}    
\begin{tabular}{llccc}
\toprule
          && \textsc{In-Distr.}   & OOD     & OOD  \\

\cmidrule{3-5}
          && $n\in$ & $n\in$ & $n\in$     \\
          && $[1,50]^*$   & $[51,100]$      & $[101,150]$  \\
\midrule
\multirow{2}{*}{$\mathtt{a}^n\mathtt{b}^n$} &Tr.$-$PE& $\textbf{100.0}_{\pm 0.0}$                   & $\textbf{100.0}_{\pm 0.0}$                   &$\text{91.3}_{\pm 8.4}$                  \\
&LSTM & $\textbf{100.0}_{\pm 0.0}$                   & $\textbf{100.0}_{\pm 0.0}$                  & $\textbf{100.0}_{\pm 0.0}$                  \\
\midrule
\multirow{2}{*}{$\mathtt{a}^n\mathtt{b}^n\mathtt{c}^n$} &Tr.$-$PE& $\textbf{100.0}_{\pm 0.0}$                    & $\textbf{100.0}_{\pm 0.0}$                  & $\text{36.0}_{\pm 14.2}$                    \\
&LSTM & $\textbf{100.0}_{\pm 0.0}$                   & $\textbf{100.0}_{\pm 0.0}$                   & $\textbf{100.0}_{\pm 0.0}$                 \\
 \midrule
 \multirow{2}{*}{$\mathtt{a}^n\mathtt{b}^n\mathtt{c}^n\mathtt{d}^n$} &Tr.$-$PE& $\textbf{100.0}_{\pm 0.0}$                    &  $\textbf{100.0}_{\pm 0.0}$                   &  $\text{24.0}_{\pm 10.2}$                 \\
 &LSTM & $\textbf{100.0}_{\pm 0.0}$                 &  $\textbf{100.0}_{\pm 0.0}$                 &  $\text{48.7}_{\pm 13.6}$                  \\
  \midrule
 \multirow{2}{*}{$\mathtt{a}^n\mathtt{b}^n\mathtt{c}^n\mathtt{d}^n\mathtt{e}^n$} &Tr.$-$PE& $\textbf{100.0}_{\pm 0.0}$                   & $\text{85.3}_{\pm 15.4}$ & $\text{3.3}_{\pm 4.7}$                   \\
 &LSTM & $\textbf{100.0}_{\pm 0.0}$                  & $\textbf{100.0}_{\pm 0.0}$                 & $\textbf{100.0}_{\pm 0.0}$                 \\

\bottomrule
\end{tabular}
\caption{\label{tab:multiple}
Multiple Agreements: Transformers without PE demonstrated the ability to extrapolate, though in general still not as good or consistent as LSTMs. $^*$The in-distribution set is held-out and only has strings with $n \in \{5,15,25,35,45\}$.
}
\end{table}

We annotate the mean and standard deviation in percentages among all attention values from one query symbol to one key symbol in Figure \ref{fig: multiple}. It can be observed that every input alphabet symbol attends to different symbols in the output alphabet using a different weight, but the attention values to each occurrence of the same symbol are similar, and thus have low variance, culminating in the grid pattern we see on the maps. The differences in weights from a query symbol suggested a particular dependency analysis learned by that run.

Unlike what we have discussed for copying and crossing, the multiple agreements strings do not have a definite dependency relation to learn, and many possible analyses exist for the same string, e.g., any two symbols in the alphabet could be pairwise dependent, or all symbols in the alphabet could be dependent on each other (cf. \citet{joshi_1985}). Thus, from run to run, to which key symbol the query symbol gives the most attention, and how much attention is given to each symbol, could indeed vary. Also, the lack of a gold analysis has determined that the model cannot simply focus on some of the symbols, and it is crucial for every symbol to attend to every symbol as we see here.

\begin{figure}
    \centering
    \includegraphics[width=\columnwidth]{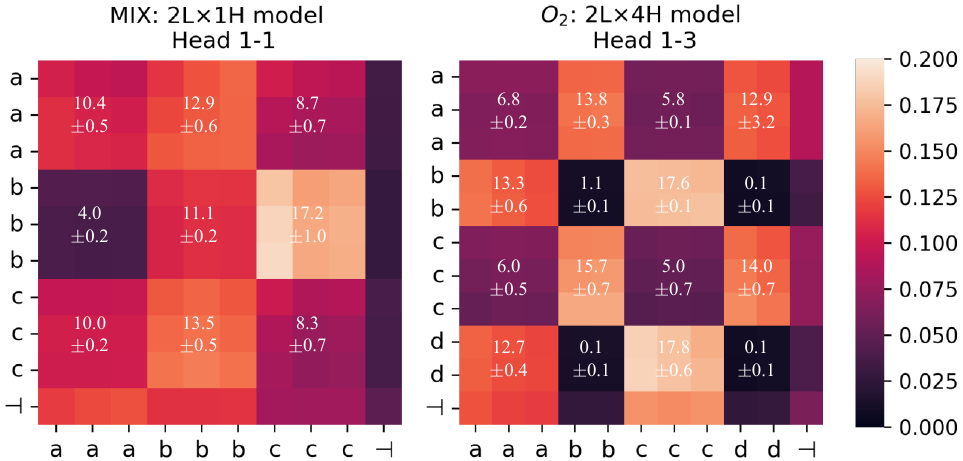}
    \caption{MIX resembles multiple agreements in that the attention weights from one query symbol to one key symbol are similar. $O_2$ resembles the checkerboard in crossing although it does not ignore non-dependents.}
    \label{fig: scramble}
\end{figure}

\subsection{Scramble Languages}

We use the macro F-1 score as the metric for this set of languages, since our data generation is skewed towards negative examples. Despite the seeming complexity of the data, Transformers are able to solve MIX and $O_2$ in-distribution test sets perfectly, while LSTMs also have very high scores. However, the MIX OOD sets are challenging for both models, while LSTMs outperformed Transformers in solving the OOD sets for $O_2$ (Table \ref{tab:scramble}).

Since $\mathtt{a}^n\mathtt{b}^n\mathtt{c}^n \subset \text{MIX}$ and $\mathtt{a}^n\mathtt{b}^m\mathtt{c}^n\mathtt{d}^m \subset O_2$, we use unscrambled strings in the visualizations for readability in Figure \ref{fig: scramble}. MIX has a pattern that resembles the one in multiple agreements in that the amount of attention from one symbol in query to one symbol in key among all occurrences is similar and has low variance, as we annotated in the visualized example. Similarly, $O_2$ has a pattern that resembles the checkerboard in crossing, although it is not the case that the queries ignore non-dependents. However, it is still evident that each query in $O_2$ identified which two symbols should form pairwise dependents and used similar attention weights to the pair. Do note that although we visualized the unscrambled strings for readability, the similar attention and low variance properties hold for other scrambled strings.

As an additional analysis, we probe the MIX representations to see what information is encoded. One possibility is the count for the symbol occurrences at each timestep, which directly follows from MIX's definition. We decode the MIX embeddings for a full counting target that maintains the ongoing tallies for all 3 symbols, as illustrated in Table \ref{tab:target}. We also include a control task \citep{hewitt-liang-2019-designing} target that is the random shuffle of the counting target, and if the probing model trained on this target has a higher error, we would be more confident that the count is actually present in the representations, rather than the counting results following from the power of the probing model.

\begin{table}
\centering
\footnotesize
\addtolength{\tabcolsep}{-2.5pt}    
\begin{tabular}{llccc}
\toprule
          && \textsc{In-Distr.}   & OOD     & OOD  \\
\midrule
          && $|w|_{\sigma}\in [1,4]$  & $|w|_{\sigma}=5$      & $|w|_{\sigma}=6$  \\
\cmidrule{3-5}
\multirow{2}{*}{MIX} 
 &Transf.& $\textbf{100.0}_{\pm 0.0}$                    &$\text{65.6}_{\pm 2.9}$                  &   $\text{45.7}_{\pm 6.3}$                    \\
 &LSTM & $\textbf{100.0}_{\pm 0.0}$                    & $\text{70.3}_{\pm 10.5}$                   & $\text{49.0}_{\pm 15.5}$             \\ 
\midrule
          && $|w|_{\sigma}\in [1,3]$  & $|w|_{\sigma}\in [1,4]$      & $|w|_{\sigma}\in [1,5]$  \\
\cmidrule{3-5}
\multirow{2}{*}{$O_2$} 
 &Transf.& $\textbf{100.0}_{\pm 0.0}$ & $\text{60.5}_{\pm 8.5}$ & $\text{45.1}_{\pm 10.1}$                    \\
 &LSTM & $\textbf{100.0}_{\pm 0.0}$                    & $\textbf{100.0}_{\pm 0.0}$                   &  $\textbf{98.6}_{\pm 0.4 }$                 \\
\bottomrule
\end{tabular}
\caption{\label{tab:scramble}
Scramble macro F-1 (\%): the models performed perfectly for in-distribution tests, but MIX OOD sets are challenging to both models, whereas LSTMs outperformed Transformers for $O_2$ OOD sets. Note that we made sure that the three tests for $O_2$ do not share datapoints when generating them.
}
\end{table}

\begin{table}
\centering
\addtolength{\tabcolsep}{+4.6pt}    
\begin{tabular}{lcccccc}
\toprule
  & \multicolumn{3}{c}{Counting Target} & \multicolumn{3}{c}{Control Task} \\
\cmidrule(lr){2-4} \cmidrule(lr){5-7} 
  & \#a        & \#b      & \#c       & \#a        & \#b      & \#c\\
$\mathtt{a}$ & {[}1       & 0        & 0{]}      & {[}2       & 2        & 2{]}      \\
$\mathtt{b}$ & {[}1       & 1        & 0{]}      & {[}1       & 0        & 0{]}      \\
$\mathtt{c}$ & {[}1       & 1        & 1{]}      & {[}2       & 1        & 1{]}      \\
$\mathtt{a}$ & {[}2       & 1        & 1{]}      & {[}1       & 1        & 0{]}      \\
$\mathtt{b}$ & {[}2       & 2        & 1{]}      & {[}1       & 1        & 1{]}      \\
$\mathtt{c}$ & {[}2       & 2        & 2{]}      & {[}2       & 2        & 1{]}      \\
\bottomrule
\end{tabular}
\caption{\label{tab:target}
Target examples for $w = \mathtt{abcabc}$. The counting target is the count of each symbol at each timestep; the control task target is the random shuffle of the counting target.
}
\end{table}

Similar to the methodology in \citet{wallace-etal-2019-nlp}, we used an MLP regressor prober with 1 hidden layer, ReLU activations, and MSE loss. We trained the prober for up to 300 epochs with early stopping. On the in-distribution test set, the prober using the counting target has an MSE of 0.21 and a Pearson correlation of 0.929 between the target and the predicted count values. This contrasts with the control task target which has an MSE of 1.33. We find this to be suggestive that the learned representations contain count information, which may have been useful for solving scramble languages.

\section{Discussion and Conclusion}

We empirically studied the Transformer's ability to learn a variety of linguistically significant {MCSLs}. The significance of the languages is two-fold: they represent a hypothesized upper bound for the complexity of natural language, and they are the abstractions of the motivating linguistic phenomena. Overall, the Transformers performed well in in-distribution tests and are comparable to LSTMs, but their ability to extrapolate is limited. In our next character prediction experiments with Transformer decoders, removing the sinusoidal PE alleviated the problem, which is an established empirical finding for some formal languages and natural language, but this technique is not always generalizable to other data, nor does it work in the encoder since the decoder can rely on the indirect positional information from the causal mask in absence of PE.

Transformers leveraged the attention mechanism to score the similarity between the substrings. In our analyses, the learned self-attention's alignments often reflect the symbol dependency relations within the string, which had been useful for MCSLs because of the rich and complex dependencies in the languages. In a more complex language like MIX, Transformers had implicitly learned some form of counting behavior that may have helped solve the language.

Within the same family of languages spanning across complexity classes, the learned patterns are similar and no significant differences in behaviors are observed in the reduced or added complexity languages. This may suggest that we cannot draw parallels between the MCSG formalisms and the Transformer's expressiveness directly, like some other formal models such as circuits \citep{hao-etal-2022-formal, merrill-etal-2022-saturated} do. However, this work serves as an example of how we may draw inspiration from the rich MCSL scholarships to motivate work in NLP, as they help us examine the linguistic capacity of current and future NLP models.

\section*{Limitations}
An empirical study on formal language learning is always inconveniently insufficient, as there is always some string length upper bound that any experiment can get to or reasonably work with, so any conclusions drawn are based on an unintentionally bounded dataset, which could weaken the argument about learnability in general as the dataset might form a language with reduced complexity.

In addition, the roles of the other heads, the feed-forward sublayer, etc. are not investigated. Therefore, we cannot definitively say how self-attention directly contributed to inference, despite learning meaningful and interpretable patterns (cf. \citet{wen2023interpretability}).

Ideally, we would complement the empirical findings with theoretical constructions on whether and how the MCSLs can be learned, which is lacking in the current work. However, the empirical results serve as the foundation towards that goal. Especially, the highly interpretable self-attention patterns could inspire us and hint at what the theoretical constructions would look like.

\bibliography{anthology,custom}
\bibliographystyle{acl_natbib}

\appendix

\section{Additional Experiment Details}
\label{sec:details}

\begin{table*}
\centering
\addtolength{\tabcolsep}{+0.6pt}
\small
\begin{tabular}{lrrrrrrrrrr}
\toprule
           & \multicolumn{2}{c}{Train}                         & \multicolumn{2}{c}{Dev}                           & \multicolumn{2}{c}{Test}                          & \multicolumn{2}{c}{OOD-1}                         & \multicolumn{2}{c}{OOD-2}                         \\
           & \multicolumn{1}{c}{\textsc{pos}} & \multicolumn{1}{c}{\textsc{neg}} & \multicolumn{1}{c}{\textsc{pos}} & \multicolumn{1}{c}{\textsc{neg}} & \multicolumn{1}{c}{\textsc{pos}} & \multicolumn{1}{c}{\textsc{neg}} & \multicolumn{1}{c}{\textsc{pos}} & \multicolumn{1}{c}{\textsc{neg}} & \multicolumn{1}{c}{\textsc{pos}} & \multicolumn{1}{c}{\textsc{neg}} \\
\midrule
    & \multicolumn{6}{c}{$|w| \in [1,11]$} & \multicolumn{2}{c}{$|w| = 12$} & \multicolumn{2}{c}{$|w| = 13$}   \\
\cmidrule(lr){2-7} \cmidrule(lr){8-9} \cmidrule(l){10-11}
$ww^\mathcal{R}$        & 2858                    & 2863                    & 609                     & 617                     & 624                     & 603                     & 4096                    & 4096                    & 8192                    & 8192                    \\
$ww$         & 2873                    & 2847                    & 622                     & 603                     & 592                     & 635                     & 4095                    & 4095                    & 8192                    & 8192                    \\
$www$        & 2894                    & 2836                    & 607                     & 621                     & 592                     & 637                     & 4096                    & 4096                    & 8192                    & 8192                    \\
\midrule
    & \multicolumn{6}{c}{$n, m \in [1,50]$} & \multicolumn{2}{c}{$n \text{ or } m \in [51,100]$} & \multicolumn{2}{c}{$n \text{ or } m \in [101,150]$}   \\
\cmidrule(lr){2-7} \cmidrule(lr){8-9} \cmidrule(l){10-11}
$\mathtt{a}^n\mathtt{b}^m\mathtt{c}^m\mathtt{d}^n$    & 1750                    & —                       & 375                     & —                       & 375                     & —                       & 7500                    & —                       & 12500                   & —                       \\
$\mathtt{a}^n\mathtt{b}^m\mathtt{c}^n\mathtt{d}^m$   & 1750                    & —                       & 375                     & —                       & 375                     & —                       & 7500                    & —                       & 12500                   & —                       \\
\midrule
    & \multicolumn{6}{c}{$n \in [1,50]$} & \multicolumn{2}{c}{$n \in [51,100]$} & \multicolumn{2}{c}{$n \in [101,150]$}   \\
\cmidrule(lr){2-7} \cmidrule(lr){8-9} \cmidrule(l){10-11}
$\mathtt{a}^n\mathtt{b}^n$       & 40                      & —                       & 5                       & —                       & 5                       & —                       & 50                      & —                       & 50                      & —                       \\
$\mathtt{a}^n\mathtt{b}^n\mathtt{c}^n$     & 40                      & —                       & 5                       & —                       & 5                       & —                       & 50                      & —                       & 50                      & —                       \\
$\mathtt{a}^n\mathtt{b}^n\mathtt{c}^n\mathtt{d}^n$   & 40                      & —                       & 5                       & —                       & 5                       & —                       & 50                      & —                       & 50                      & —                       \\
$\mathtt{a}^n\mathtt{b}^n\mathtt{c}^n\mathtt{d}^n\mathtt{e}^n$ & 40                      & —                       & 5                       & —                       & 5                       & —                       & 50                      & —                       & 50                      & —                       \\
\midrule
    & \multicolumn{6}{c}{$|w|_\sigma \in [1,4]$} & \multicolumn{2}{c}{$|w|_\sigma = 5$} & \multicolumn{2}{c}{$|w|_\sigma = 6$} \\
\cmidrule(lr){2-7} \cmidrule(lr){8-9} \cmidrule(l){10-11}
MIX        & 25497                   & 51645                   & 5462                    & 11034                   & 5467                    & 11145                   & 756756                        & 408111                        & 756756                        & 960501                        \\
\midrule
    & \multicolumn{6}{c}{$|w|_\sigma \in [1,3]$} & \multicolumn{2}{c}{$|w|_\sigma \in [1,4]$} & \multicolumn{2}{c}{$|w|_\sigma \in [1,5]$}   \\
\cmidrule(lr){2-7} \cmidrule(lr){8-9} \cmidrule(l){10-11}
$O_2$         & 56518                   & 289590                  & 12111                   & 62006                   & 12115                   & 62156                   & 44100                        & 1135444                        & 56700                        & 3030880                       \\
\bottomrule
\end{tabular}
\caption{\label{tab:datastat}
Dataset statistics and label distribution. The next character prediction task uses only positive examples.
 OOD sets for scramble languages are downsampled. Negative examples for scramble languages also include strings where $|w|_\sigma = 0$.}
\end{table*}

\paragraph{Choice of Task}
In pilot experiments, training crossing and multiple agreements with the binary classification task was unsuccessful, and our analyses suggest that they learned spurious statistical cues. The task was especially difficult for multiple agreements, where only one example is available for each $n$. For crossing, we tried to use an equal number of positive and negative examples, but that is not enough for the model to rule out alternative wrong hypotheses. On the other hand, if we follow what we did for $O_2$ and enumerate all possible negative examples in a length range, since crossing has a much wider length range than $O_2$, this will lead to an explosion of negative examples and is impractical to work with. Teaching these two sets of languages with the binary classification setup may still be possible, but the negative examples likely need to be carefully curated, so that we avoid an explosion of negative examples over positive examples, but still have enough negative datapoints to help the model eliminate most wrong hypotheses such as the spurious cue ones.

\paragraph{The \texttt{[EOS]} Decision}
In the binary classification task, since we are not scoring or generating a string, the decision on whether to add \texttt{[EOS]} to strings is arbitrary. \citet{newman-etal-2020-eos} suggest that without \texttt{[EOS]}, models may extrapolate better to longer strings. We tried the setup without \texttt{[EOS]} in pilot experiments but found no significantly better performance for our Transformer models in the studied languages. We chose to include \texttt{[EOS]} and use the \texttt{[EOS]} embeddings as the sentence representations for LSTMs in this task.

\section{Training Details}
\label{sec:training}

\begin{table}
\centering
\footnotesize
\addtolength{\tabcolsep}{-4.3pt}    
\begin{tabular}{lllll}
\toprule
                 & Learning                   & Min. & Max. & Patience \\
                 & Rate &Delta &Epochs &\\
\midrule
Copying       & \{1e-3, 5e-4, 1e-4\} & 1e-4       & 200       & 20         \\
Crossing-LSTM & \{1e-2, 5e-2, 1e-3\} & 1e-4       & 150       & 20         \\
Crossing-Transf. & \{1e-3, 5e-4, 1e-4\} & 1e-4       & 150       & 20         \\
Multiple-LSTM    & \{1e-2, 5e-2, 1e-3\} & 1e-4       & 3000       & 400         \\
Multiple-Transf. & \{1e-3, 5e-4\}       & 1e-4       & 3000      & 400        \\
MIX              & \{1e-3, 5e-4, 1e-4\}       & 1e-5       & 50       & 5           \\
$O_2$             & \{5e-4, 3e-4, 1e-4\} & 1e-5       & 50       & 5          \\
\bottomrule
\end{tabular}
\caption{\label{tab:searchspace}
Search space and choices for certain hyperparameters per language.
}
\end{table}

The datasets for all languages are generated exactly once, and are used across hyperparameter tuning and the final experiments. We record the random seeds used for generation for reproducibility. We try to enumerate all examples in our chosen length range except for the OOD sets of the scramble languages because of an explosion of datapoints as strings get longer, and we downsample in this case. For MIX, positive examples are capped at 756756; negative examples for a given string length are capped at 25200. For $O_2$, examples for a given string length are capped at 6300.

We then do a 70\%-15\%-15\% Train-Dev-Test random split of the data. This is performed over the entire dataset, except for scramble languages, in which we split by each string length separately since the dataset is skewed towards longer strings as a result of enumerating all distinct permutations for a given string length. For multiple agreements languages, since the dataset is too small to be randomly split, we manually picked strings with $n \in \{5, 15, 25, 35, 45\}$ as the test set and strings with $n \in \{6,16,26,36,46\}$ as the development set. We give detailed statistics of the dataset size and the label distribution in Table \ref{tab:datastat}.

We first tune the hyperparameters for each of the languages on both the Transformer and the LSTM. All models tune $d_\text{model}$ in $\{16, 32, 64\}$, and for Transformers, we also tune the number of layers in $\{1,2\}$ and the number of heads in $\{1,2,4\}$. The search space for the learning rate per language is in Table \ref{tab:searchspace}. We tuned the hyperparameters using grid search and picked the configuration with the lowest development loss for each language. The final set of hyperparameters is in Table \ref{tab:hyperparams}. For both tuning and training, we also use early stopping with manually selected minimum delta, patience, and the maximum number of epochs as shown in Table \ref{tab:searchspace}. All experiment runs used a batch size of 32, and the AdamW optimizer \citep{loshchilov2018decoupled}. The embedding layer of the Transformer is initialized using a uniform distribution in the range $[-0.1, 0.1]$.

\begin{table}
\centering
\footnotesize
\addtolength{\tabcolsep}{-2.7pt}    
\begin{tabular}{lcccccc}
\toprule
           & \multicolumn{2}{c}{LSTM} & \multicolumn{4}{c}{Transformer} \\
\cmidrule(l){2-3} \cmidrule(l){4-7}
           & $d_\text{model}$      & LR & $d_\text{model}$ & LR &\#layers & \#heads \\
$ww^\mathcal{R}$        & 64            & 1e-4    & 64               & 5e-4       & 2             & 4              \\
$ww$         & 64            & 5e-4    & 32               & 5e-4       & 2             & 1              \\
$www$        & 64            & 1e-4    & 64               & 5e-4       & 2             & 4              \\
$\mathtt{a}^n\mathtt{b}^m\mathtt{c}^m\mathtt{d}^n$   & 64            & 5e-2    & 64               & 1e-3       & 1             & 4              \\
$\mathtt{a}^n\mathtt{b}^m\mathtt{c}^n\mathtt{d}^m$   & 64            & 1e-2    & 64               & 1e-3       & 1             & 4              \\
$\mathtt{a}^n\mathtt{b}^n$       & 64            & 1e-2    & 64               & 5e-4       & 2             & 4              \\
$\mathtt{a}^n\mathtt{b}^n\mathtt{c}^n$     & 64            & 1e-2    & 64               & 5e-4       & 2             & 4              \\
$\mathtt{a}^n\mathtt{b}^n\mathtt{c}^n\mathtt{d}^n$   & 64            & 1e-2    & 64               & 5e-4       & 1             & 4              \\
$\mathtt{a}^n\mathtt{b}^n\mathtt{c}^n\mathtt{d}^n\mathtt{e}^n$ & 64            & 5e-2    & 64               & 5e-4       & 1             & 4              \\
MIX        & 64            & 5e-4    & 64               & 1e-4       & 2             & 1              \\
$O_2$       & 64            & 5e-4    & 64               & 5e-4       & 2             & 4             \\
\bottomrule
\end{tabular}
\caption{\label{tab:hyperparams}
Final hyperparameters used in experiments per language.
}
\end{table}

\end{document}